\documentclass[a4paper,twoside]{article}
\usepackage{epsfig}
\usepackage{subcaption}
\usepackage{calc}
\usepackage{amssymb}
\usepackage{amstext}
\usepackage{amsmath}
\usepackage{amsthm}
\usepackage{multicol}
\usepackage{pslatex}
\usepackage{apalike}
\usepackage{bm}
\usepackage[bottom]{footmisc}
\usepackage{SCITEPRESS}     

\begin{document}

\title{Deep Learning for Diagonal Earlobe Crease Detection}
\author{\authorname{Sara L. Almonacid-Uribe\orcidAuthor{0000-0001-6660-0867},
Oliverio J. Santana\orcidAuthor{0000-0001-7511-5783},\\
Daniel Hernández-Sosa\orcidAuthor{0000-0003-3022-7698}, and
David Freire-Obregón\orcidAuthor{0000-0003-2378-4277}}
\affiliation{SIANI, Universidad de Las Palmas de Gran Canaria, Las Palmas de Gran Canaria, Spain}
\email{david.freire@ulpgc.es}
}

\keywords{Computer vision, diagonal earlobe crease, DELC, Frank's Sign, cardiovascular disease, coronary artery disease, deep learning.}

\abstract{
An article published on Medical News Today in June 2022 presented a fundamental question in its title: Can an earlobe crease predict heart attacks? The author explained that end arteries supply the heart and ears. In other words, if they lose blood supply, no other arteries can take over, resulting in tissue damage. Consequently, some earlobes have a diagonal crease, line, or deep fold that resembles a wrinkle. In this paper, we take a step toward detecting this specific marker, commonly known as DELC or Frank's Sign. For this reason, we have made the first DELC dataset available to the public. In addition, we have investigated the performance of numerous cutting-edge backbones on annotated photos. Experimentally, we demonstrate that it is possible to solve this challenge by combining pre-trained encoders with a customized classifier to achieve 97.7\% accuracy. Moreover, we have analyzed the backbone trade-off between performance and size, estimating MobileNet as the most promising encoder.}

\onecolumn \maketitle \normalsize \setcounter{footnote}{0} \vfill



\section{Introduction}
\label{sec:introduction}

According to the Centers for Disease Control and Prevention (CDC), heart disease is the leading cause of death for men, women, and the majority of racial and ethnic groups in the United States \cite{CDC22}. Overall, cardiovascular disease is responsible for one death every 34 seconds in the United States. Furthermore, one in five heart attacks are silent; the damage is done, but the individual is unaware \cite{Tsao22}. Early detection is essential for providing treatment to alleviate symptoms, reduce mortality, and enhance the quality of life \cite{Boudoulas16}. 

\begin{figure}[t]  
\begin{minipage}[t]{0.9\linewidth}
    \centering
    \includegraphics[scale=0.54]{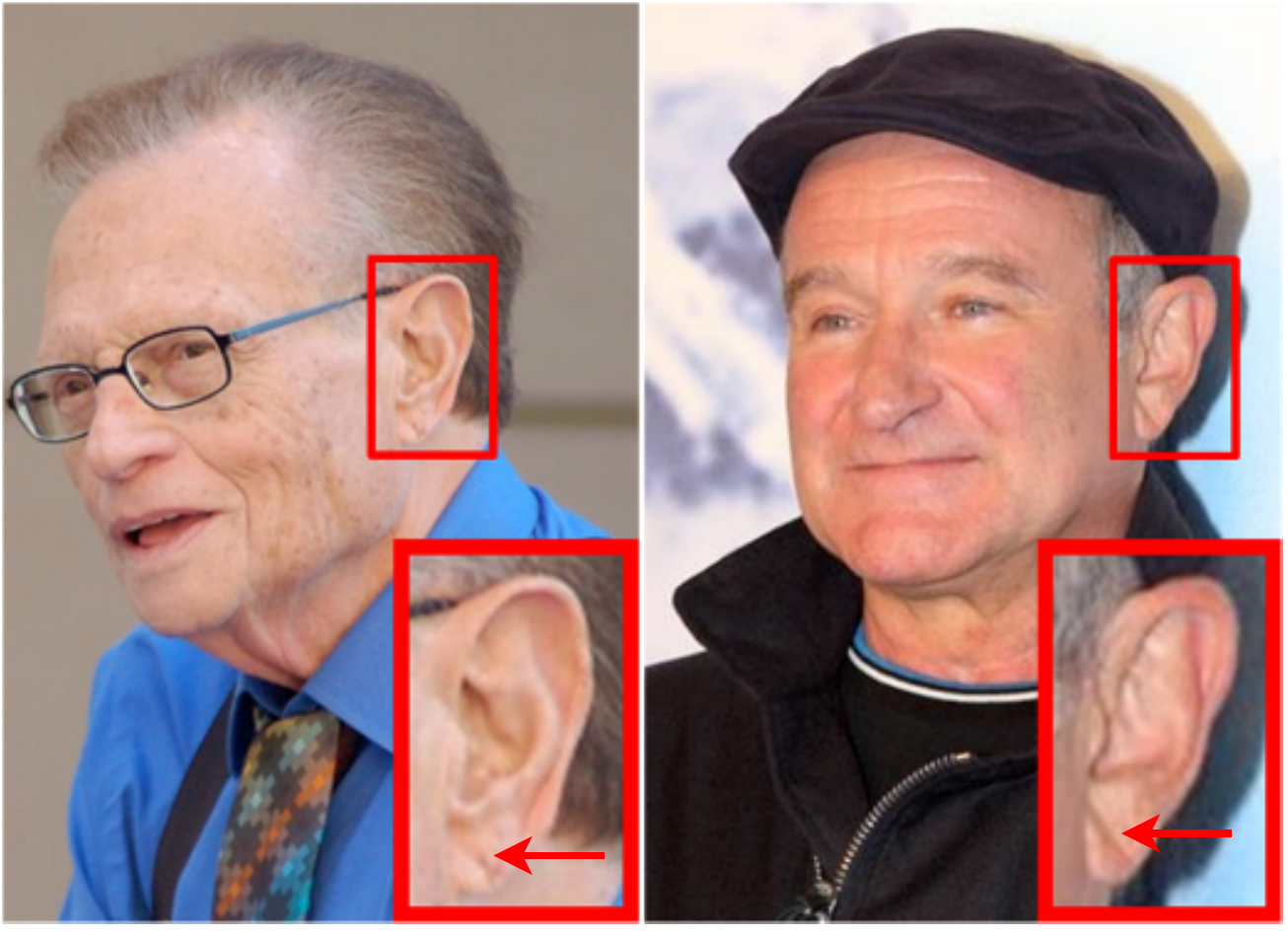}
      \end{minipage}
    \caption{\textbf{Celebrities exhibiting a DELC marker.} In 1987, the former CNN interviewer Larry King suffered a heart attack and underwent bypass surgery (photo by Eva Rinaldi, Wikimedia Commons, CC-BY-SA 2.0). In 2009, the former comedian and actor Robin Williams underwent aortic valve replacement surgery (photo by Angela George, Wikimedia Commons, CC-BY 3.0). The ear is highlighted in both pictures.
     \label{fig:intro_img}}
\end{figure}

As a standard practice, clinicians are taught to diagnose coronary artery disease (CAD) based on the medical history, biomarkers, raw scores, and physical examinations of individual patients, which they interpret based on their clinical experience. However, this approach has evolved due to technological advances. In the past decade, deep learning (DL) has demonstrated a promising ability to detect abnormalities in computed tomography (CT) images \cite{Ardila19}. Several DL techniques have been proposed to automatically estimate CAD markers from CT images. The majority of these models predict clinically relevant image features from cardiac CT, such as coronary artery calcification scoring \cite{Isgum12,Wolterink15,Zeleznik2021}, non-calcified atherosclerotic plaque localization \cite{Yamak14,Zhao19}, and stenosis from cardiac CT \cite{Lee19,Zreik19}.

Even though the development of DL on CT images is promising, CT equipment is expensive and cardiac illnesses are hard to find unless the patient has symptoms and goes to the hospital for a cardiac checkup. In this context, the diagonal earlobe crease (DELC) can be a helpful guide to identify cardiac problems. This crease extends diagonally from the tragus to the earlobe's border (see Figure \ref{fig:intro_img}).
It is also known as Frank's Sign because it was first described by Frank in a case series of CAD patients \cite{Frank73}. Since then, numerous reports have been published concerning its association primarily with atherosclerosis, particularly CAD \cite{Wieckowski21}. While not as well known as more traditional approaches, DELC examinations are painless, non-invasive, and simple to interpret. If its diagnostic accuracy is sufficient for decision-making, it could be utilized in primary care or emergency departments.

In this work, we have created a DELC detector using state-of-the-art (SOTA) backbones and ear collections as benchmarks for the models. First, we have gathered DELC ear images available on the Internet. Then, we developed multiple DL models considering pre-trained encoders, also known as backbones, to predict whether or not an ear displays a DELC.
In addition, we analyzed the performance of the considered backbones by varying the classifier parameters and found no correlation between the number of parameters and the best model.

Our proposal was evaluated using a mixed-source dataset. As previously stated, we gathered 342 positive DELC images by collecting publicly available images from the Internet and cropping off the ears. Negative samples were obtained from a publicly accessible ear database, namely AWE \cite{Emersic17}. All images are collected from natural settings, including lighting, pose, and shape variations. Considering the number of samples, data augmentation techniques were considered during training. The outcomes are remarkable (predictions up to 97.7\% accurate) and have yielded intriguing insights. 

\begin{figure}[t]  
\begin{minipage}[t]{0.9\linewidth}
    \centering
    \includegraphics[scale=0.7]{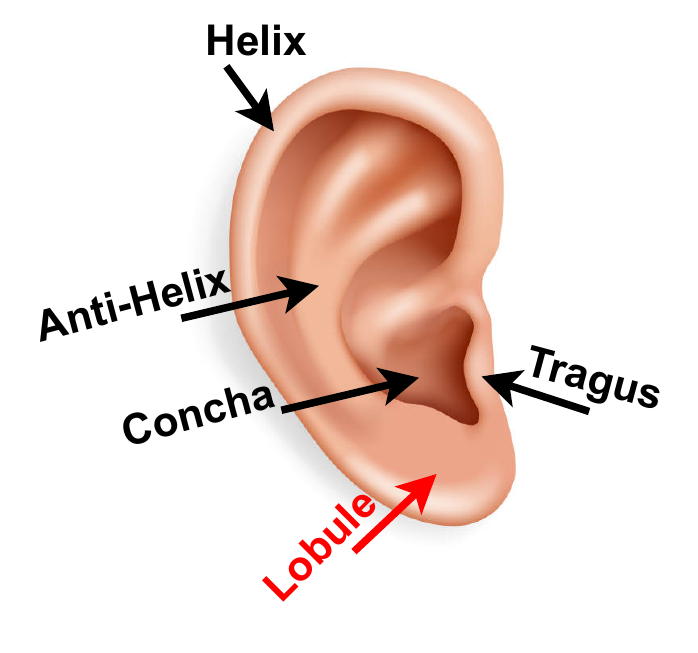}
      \end{minipage}
    \caption{\textbf{Outer ear scheme.}  The human earlobe (lobulus auriculae) is composed of areolar and adipose connective tissues that lack rigidity. Due to the absence of cartilage in the earlobe, it has an abundant blood flow and may aid in warming the ears \cite{Steinberg03}. 
     \label{fig:earlobe}}
\end{figure}

Even though the earlobe is a relatively small part of the ear (see Figure \ref{fig:earlobe}), the classifier's performance is noteworthy. Unlike other diseases, such as melanoma, which can be found anywhere in the human body, the DELC is located in a specific area, facilitating the detection task. The usability of the trained models reveals an additional insightful revelation. Light-weight convolutional neural networks such as MobileNet provide high accuracy, balancing the precision of complex neural network structures with the performance constraints of mobile runtimes. Hence, ubiquitous applications could take advantage of this proposal, making it possible to detect DELC by just using a smartphone anywhere and anytime. Our contributions can be summarized as follows:

\begin{itemize}
    \item We propose a novel dataset with DELC ear images in the wild with 342 samples. All samples have been gathered from the Internet. The dataset is publicly available.
    \item We experimentally demonstrate that it is possible to tackle this problem by combining pre-trained backbones with a new classifier. 
    \item In this experiment, eleven different backbones are compared to one another regarding their DELC detection performance. Moreover, the models' size-performance trade-off analysis demonstrates that the problem can be effectively addressed by employing light-weight encoders. As aforementioned, this opens the door for the broad implementation of this technology.
\end{itemize}

The remainder of this paper is organized as follows. Section~\ref{sec:relatedwork} discusses previous related work. Section~\ref{sec:pipeline} describes the proposed pipeline. Section~\ref{sec:results} reports the experimental setup and the experimental results. Finally, conclusions are drawn in Section~\ref{sec:con}.

\section{Related work}
\label{sec:relatedwork}

The state of the art can be studied from both physiological and technological viewpoints. The former aims to find support for the relationship between CAD and DELC by examining related studies, while the latter intends to evaluate the Computer Vision Community proposals.

\subsection{Physiological relevance}
Several investigations over the previous few decades have established an association between DELC and cardiac issues. DELC is a unilateral or bilateral diagonal fold in the earlobe, typically making a 45-degree angle from the intertragic notch to the posterior edge of the ear. This marker has a grading system linked to the incidence of CAD based on numerous characteristics, including length, depth, bilateralism, and inclination. Complete bilateralism is regarded as the most severe condition \cite{Rodri15}.

As previously stated, Frank established the initial idea for this association \cite{Frank73}. According to some scientists, it indicates physiological aging \cite{Mallinson17}. Nonetheless, the CAD concept began to gather support, and several additional researchers conducted experiments demonstrating that this link can accurately predict if a patient is prone to cardiovascular issues.  It should be noted that coronary disease is one of the leading causes of death in developed nations \cite{Sanchis16}; hence, early detection is essential for enhancing the patient's quality of life and preventing or lowering CAD-related mortality. 

A pioneer work evaluated 340 patients, of whom 257 had CAD \cite{Pasternac82}. It was determined that 91\% of patients with DELC had CAD, the most prevalent sign in those with more severe disease. More recently, Stoyanov et al. investigated 45 patients, 16 females and 29 males \cite{Stoyanov21}. Twenty-two individuals had a well-documented clinical history of CAD, while the remaining patients did not. Upon general examination before the autopsy, 35 patients had well-formed DELC. In addition, patients with pierced ears had no signs of lobule injury due to piercing. Hence the observed creases were accepted as DELC.

\subsection{Computer vision relevance}

For decades, biometric traits have been explored in Computer Vision. Recent research has focused on gait analysis or body components to address a variety of tasks, including violence detection \cite{freire22mvap}, facial expressions \cite{ojsantana22mtool}, face/voice verification \cite{freire21prl}, and forensics \cite{castrillon18prl}. Ear recognition has also been widely addressed \cite{Galdamez17,Alshazly19}. 

For healthcare use, relevant proposals diagnose several ear-related diseases such as otitis media, attic retraction, atelectasis, tumors, or otitis externa. These categories encompass most ear illnesses diagnosed by observing the eardrum with an otoendoscopy \cite{Dongchul19}. Recently, Zeng et al. combined several pre-trained encoders to achieve a 95,59\% accuracy on detecting some of these illnesses using otoendoscopy images as input \cite{Zeng21}. These authors argued that using pre-trained DL architectures provides a remarkable advantage over traditional handcrafted methods. To diagnose Chronic Otitis, Wang et al. proposed a deep-learning system that automatically retrieved the region of interest from two-dimensional CT scans of the temporal bone \cite{Wang20}. These authors asserted that their model's performance (83,3\% sensitivity and 91.4\% specificity) was equivalent and, in some instances, superior to that of clinical specialists (81,1\% sensitivity and 88.8\% specificity). We have also adopted a DL approach to tackle the DELC detection problem. In contrast, we aim to use the ear as a marker for CAD.

Hirano et al. published an experimental study analyzing DELC and CAD \cite{Hirano16}. Their research employed a handcrafted approach (manually trimmed earlobes and a Canny edge detector) to detect DELC in meticulously captured images. For this experiment, 88 participants' ears were photographed from a single frontal angle. Only 16\% of the participants were healthy. Unlike this study, we considered images of ears in the wild, which varied greatly in pose and illumination.

\begin{figure}[t]  
\begin{minipage}[t]{0.9\linewidth}
    \centering
    \includegraphics[scale=0.7]{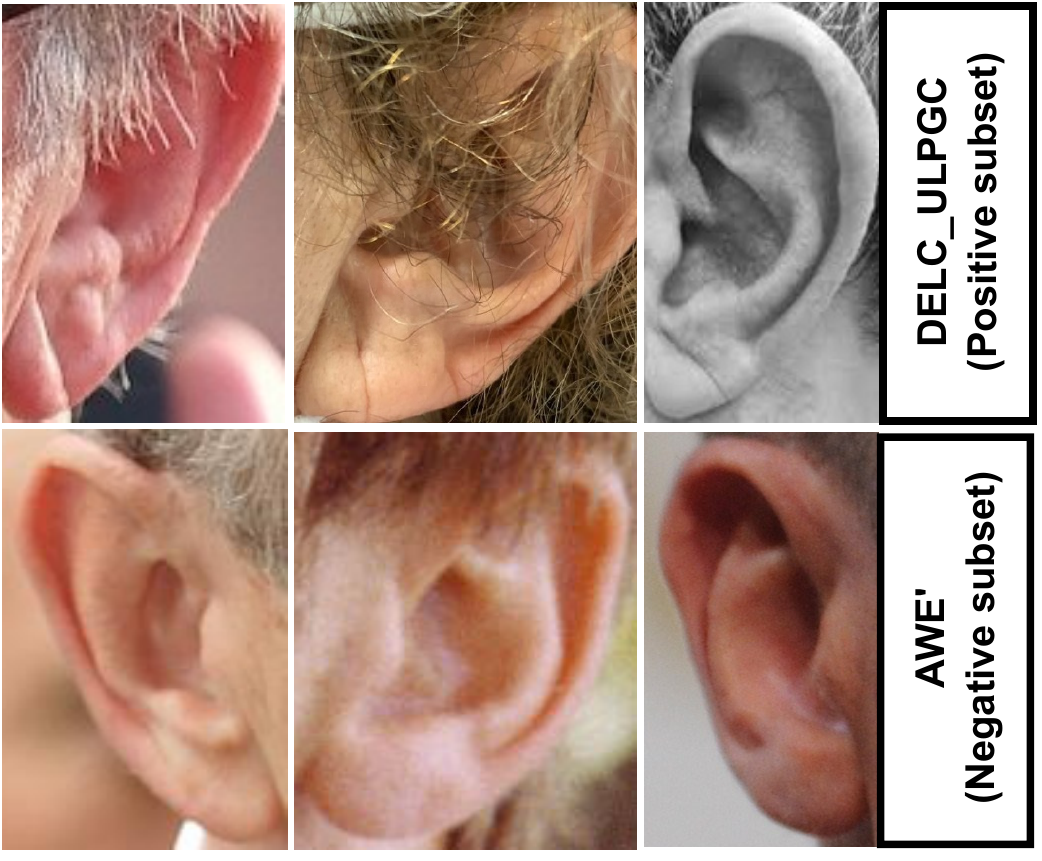}
      \end{minipage}
    \caption{\textbf{DELC\_ULPGC+AWE' Dataset.} The studied dataset comprises two in-the-wild subsets. Both of them are gathered from the Internet: a subset of the well-known AWE dataset \cite{Emersic17} as the DELC negative subset and the new proposed DELC\_ULPGC subset. 
     \label{fig:dataset}}
\end{figure}

\begin{figure*}[!t]
\centering
  \includegraphics[scale=0.75]{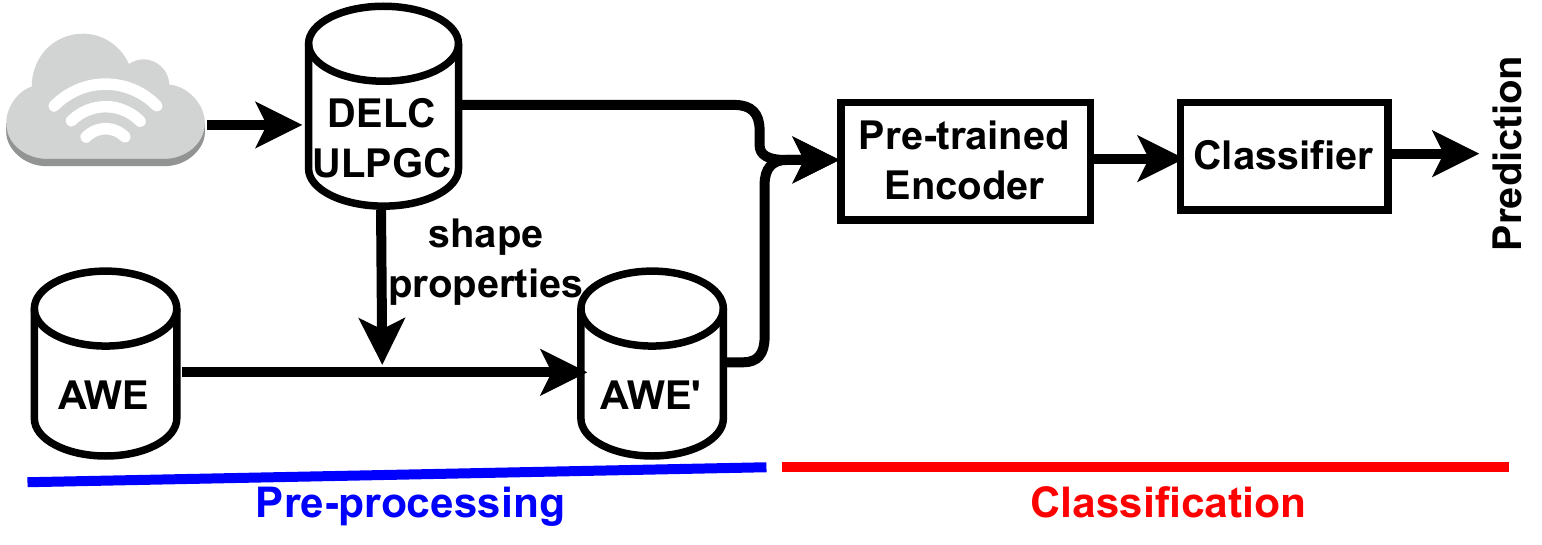}
    \caption{\textbf{The proposed pipeline for the DELC detection system.} The devised process comprises two main modules: the ear pre-processing module and the classification module. In the first module, the DELC\_ULPGC+AWE' dataset is generated and passed to the second module, where features are computed. The resulting tensor acts as an input to the classifier, completing the process.}
\label{fig:pipeline}
\end{figure*}

\section{Description of the proposal}
\label{sec:pipeline}
This paper presents and assesses a sequential training pipeline consisting of two primary modules: a module for the production of datasets and a module implementing a pre-trained backbone followed by a trainable classifier that computes a distance measure of the generated embeddings.
This distance metric is used to calculate a loss function that is sent back to the classifier. Figure \ref{fig:pipeline} shows a schematic representation of the approach described in this paper.

\subsection{DELC\_ULPGC+AWE' Dataset}
To our knowledge, there is no DELC dataset available to the public.
The collection procedure relies on the idea of generability.
We intend to develop robust detection models for use in the field.
This collection of pictures was acquired from the Internet under unrestricted conditions. Consequently, images exhibit substantial differences in the pose, scale, and illumination (see Figure \ref{fig:dataset}). The procedure involves four steps:
\begin{enumerate}
\item Web scraping. We conducted a search using key terms such as ``DELC", ``Frank's Sign", and several celebrity names to download images.
\item The labeling phase of ears includes trimming the ear region. The selected tool to perform this task is labelImg\footnote{https://github.com/heartexlabs/labelImg}.
We did not differentiate ear sections; the entire ear was cut off.
Unlike earlier handcrafted techniques \cite{Hirano16}, DL algorithms can detect across a larger region. After the second stage, the subset of 342 positive DELC images is available.
\item We compute statistical shape properties (mean and standard deviation) of the positive subset. This information is later used to obtain the negative DELC subset. The mean shape of the positive subset is $82\times159$.
\item We have considered the publicly available AWE dataset to generate the negative subset \cite{Emersic17}. The AWE data collection introduces the concept of \textit{ear images captured in the wild}. Emersic et al. collected Internet-based celebrity photos. Each subject comprises ten photos, ranging in size from $15\times29$ pixels for the smallest sample to $473\times1022$ pixels for the largest. We selected this dataset due to the nature of the gathering process, which is identical to ours. To this end, we extracted images within the resolution of the DELC-positive subset. Finally, to ensure no DELC-positive samples were in the selected images, we examined them to generate a 350-image subset of negative images, namely AWE' subset. 
\end{enumerate}

\subsection{The proposed architecture}

The implemented encoding transforms the input data into a vector of features. Initially, each input sample is sent to the encoder that has been trained to extract features. These encoders are trained on the ImageNet dataset with 1000 distinct classes \cite{Deng09}.
Convolutional layers closest to the encoder's input layer learn low-level features such as lines, whereas layers in the middle of the encoder learn complicated abstract characteristics. The last layers interpret the retrieved features within the context of a classification task.

The trainable classifier refines and condenses the previously computed features into a smaller, more specific set of features. It consists of two dense layers, each with 1024 units. Finally, a sigmoid activation function generates the classification output. 

\subsection{The adopted experimental protocol}
\textbf{Data augmentation}. The collection under consideration contains an insufficient number of samples. For instance, nearly 350 samples per class are inadequate to train a classifier without overfitting. A strategy for augmenting data yielded 2100 photos per class. The transformations utilized for data augmentation include random brightness, random contrast, motion blur, horizontal flip, shift, scale, and rotate \cite{Buslaev20}. Augmented subsets are exclusively utilized for training purposes. 

\textbf{Backbone comparison}. Several pre-trained encoders were compared: VGGNet \cite{Simonyan15}, InceptionV3 \cite{Szegedy16}, ResNet \cite{He16}, Xception \cite{Chollet16}, MobileNet \cite{Howard17} and DenseNet \cite{Huang16}. As aforementioned, these backbones were trained using the ImageNet dataset with 1000 distinct classes \cite{Deng09}. The pipeline depicted in Figure \ref{fig:pipeline} paired the considered pre-trained encoder with a trainable classifier, utilizing the Adam optimizer \cite{Kingma15} with a learning rate of $10^{-3}$ and a decay rate of $0.4$.

\section{Experimental setup}
\label{sec:results}
This section is divided into two subsections related to the setup and results of the designed experiments. The first subsection describes the technical details of our proposal, such as the loss function and the data split. The achieved results are summarized in the second subsection.

\subsection{Experimental setup}

As mentioned above, the classical detection scenario in classification aims to determine which class belongs to a sample. In this regard, we have considered two classes: DELC and not DELC. Since it is a binary classifier, the considered loss function to tackle the problem is the binary cross-entropy:

\begin{equation}
    Loss = -1/N * \sum_{i=1}^{N}-{(y_i\log(p_i) + (1 - y_i)\log(1 - p_i))}
\end{equation}

Where $p_i$ is the i-th scalar value in the model output, $y_i$ is the corresponding target value, and N is the number of scalar values in the model output.

Finally, the results presented in this section refer to the average accuracy on five repetitions of 9-fold cross-validation. On average, 615 original samples are selected for training, and the remaining 69 samples are used for the test. Contrary to test samples, the selected training samples are augmented during training.

\begin{table}[ht]
\caption{\textbf{Absolute comparison of different backbones on the DELC\_ULPGC+AWE' dataset}. The table is organized in terms of backbone, validation accuracy (Val. Acc.), test accuracy (Test Acc.), and the number of parameters of the backbone (\#$B_{Param}$). The bold entries show the best result and the lightest model.}\label{tab:res} \centering
\begin{tabular}{|l|c|c|r|}
  Backbone & Val. Acc. & Test Acc. & \#$B_{Param}$ \\
  \hline
  Xception & $95,1$\% & $94,1$\% & 22.9M\\
  \hline
  VGG16 & $96,5$\% & $93,9$\% & 138.4M\\
  \hline
  VGG19 & $95,1$\% & $92,7$\% & 143.7M\\
  \hline
  ResNet50 & $98,1$\% & $95,8$\% & 25.6M\\
  \hline
  ResNet101 & $97,5$\% & $94,8$\% & 44.7M\\
  \hline
  ResNet152 & $97,8$\% & $95,1$\% & 60.4M\\
  \hline
  MobileNet & $98,7$\% & $96,7$\% & \textbf{4.3M}\\
  \hline
  InceptionV3 & $98,9$\% & $\bm{97,7}$\% & 23.9M\\
  \hline
  DenseNet121 & $96,4$\% & $95,5$\% & 8.1M\\
  \hline
  DenseNet169 & $88,7$\% & $88,1$\% & 14.3M\\
  \hline
  DenseNet201 & $95,1$\% & $93,4$\% & 20.2M\\
  \hline
\end{tabular}
\end{table}

\begin{figure}[t]  
\begin{minipage}[t]{0.9\linewidth}
    \centering
    \includegraphics[scale=0.47]{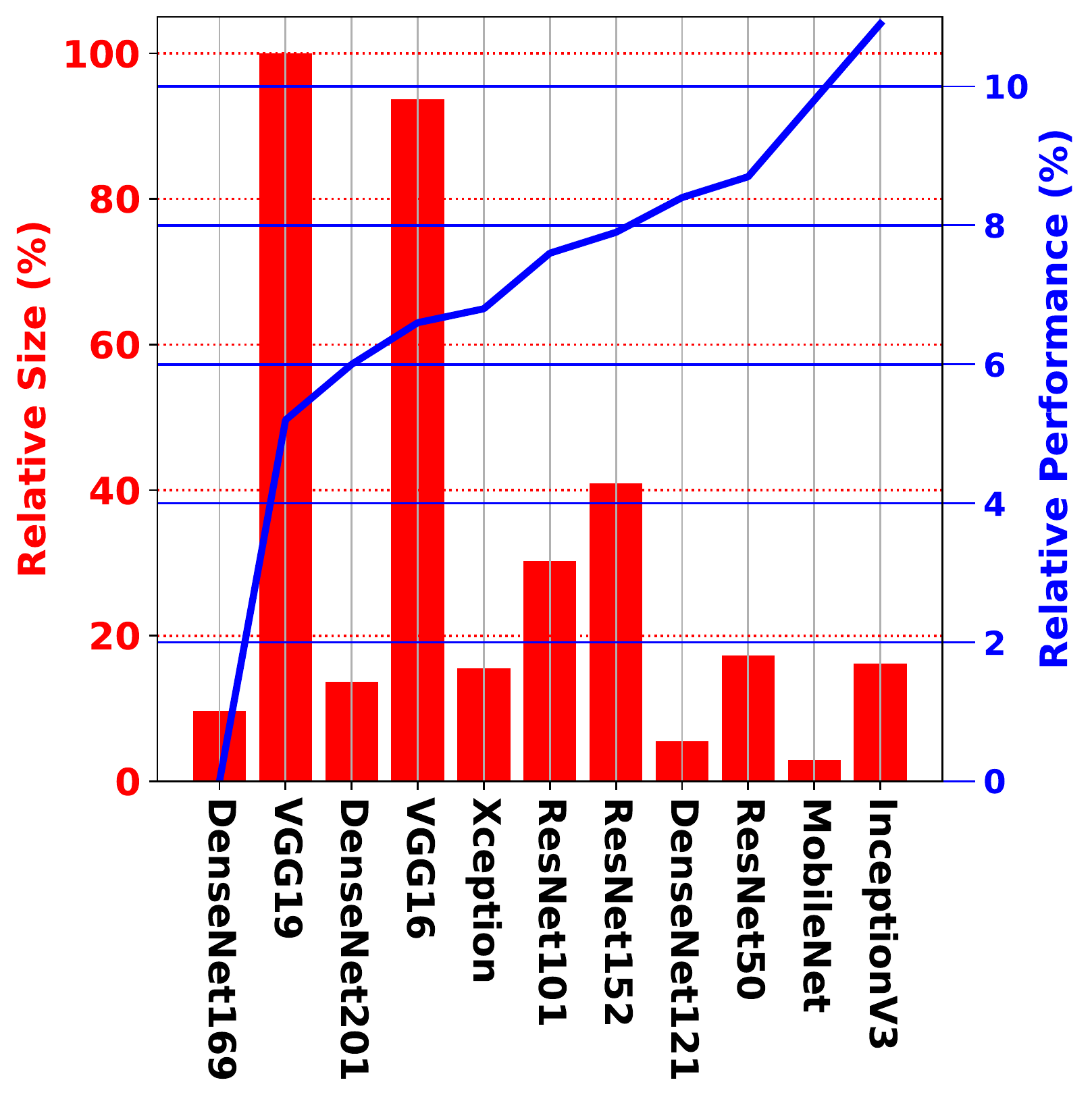}
      \end{minipage}
    \caption{\textbf{Relative comparison performance and size of encoders.} The blue line stands for the relative performance between encoders, whereas the red bars stand for the relative size of the encoder. The higher the blue line, the better, meaning a higher performance. The smaller the red bar, the better, meaning the model is lighter than others with a higher bar.
     \label{fig:relperf}}
\end{figure}
 
\subsection{Results}

\begin{figure*}[!htb]
   \begin{minipage}{0.48\textwidth}
     \centering
     \includegraphics[width=.8\linewidth]{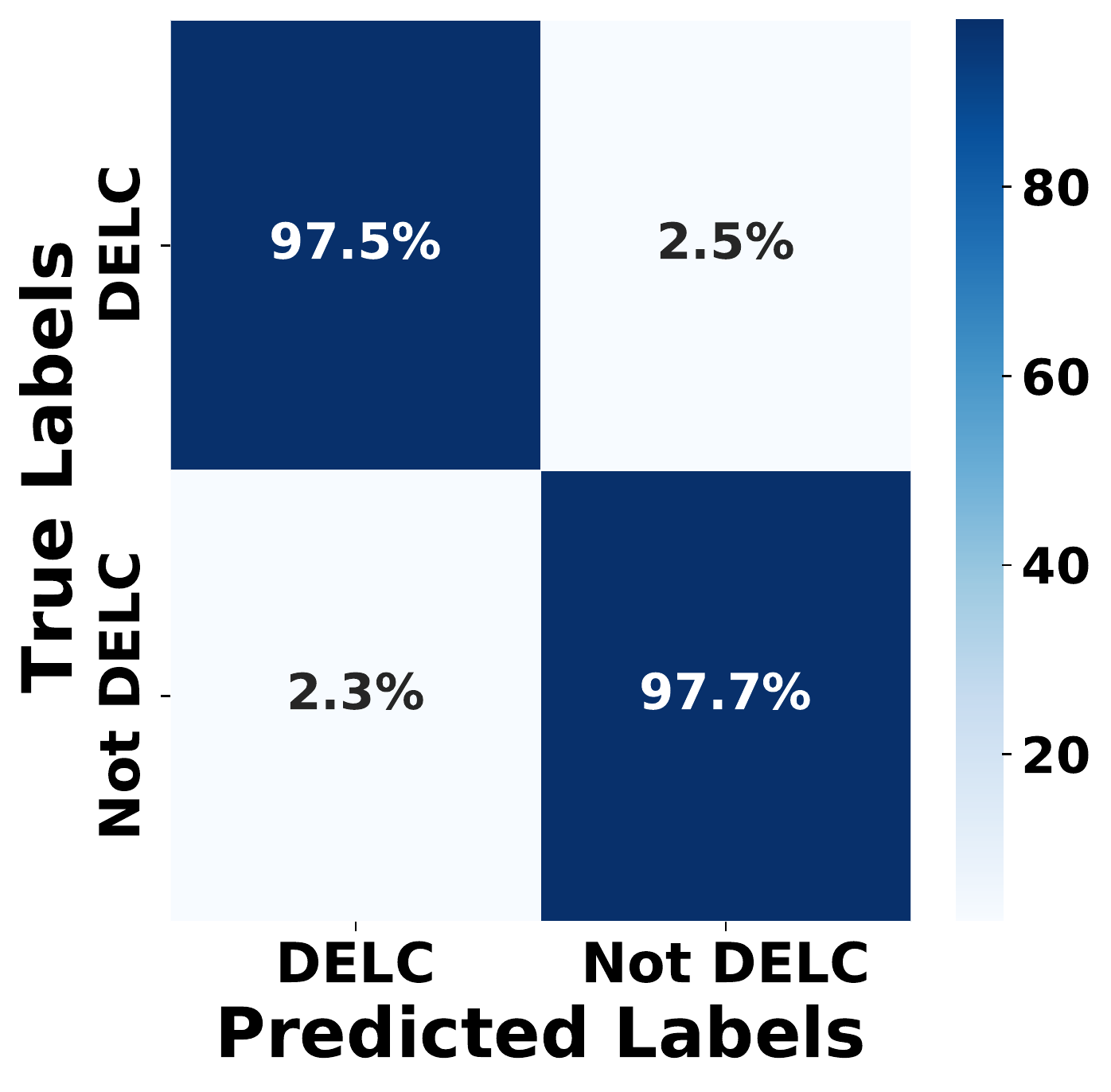}
     \caption{\textbf{InceptionV3 Confusion Matrix.}}\label{Fig:CF1}
   \end{minipage}\hfill
   \begin{minipage}{0.48\textwidth}
     \centering
     \includegraphics[width=.8\linewidth]{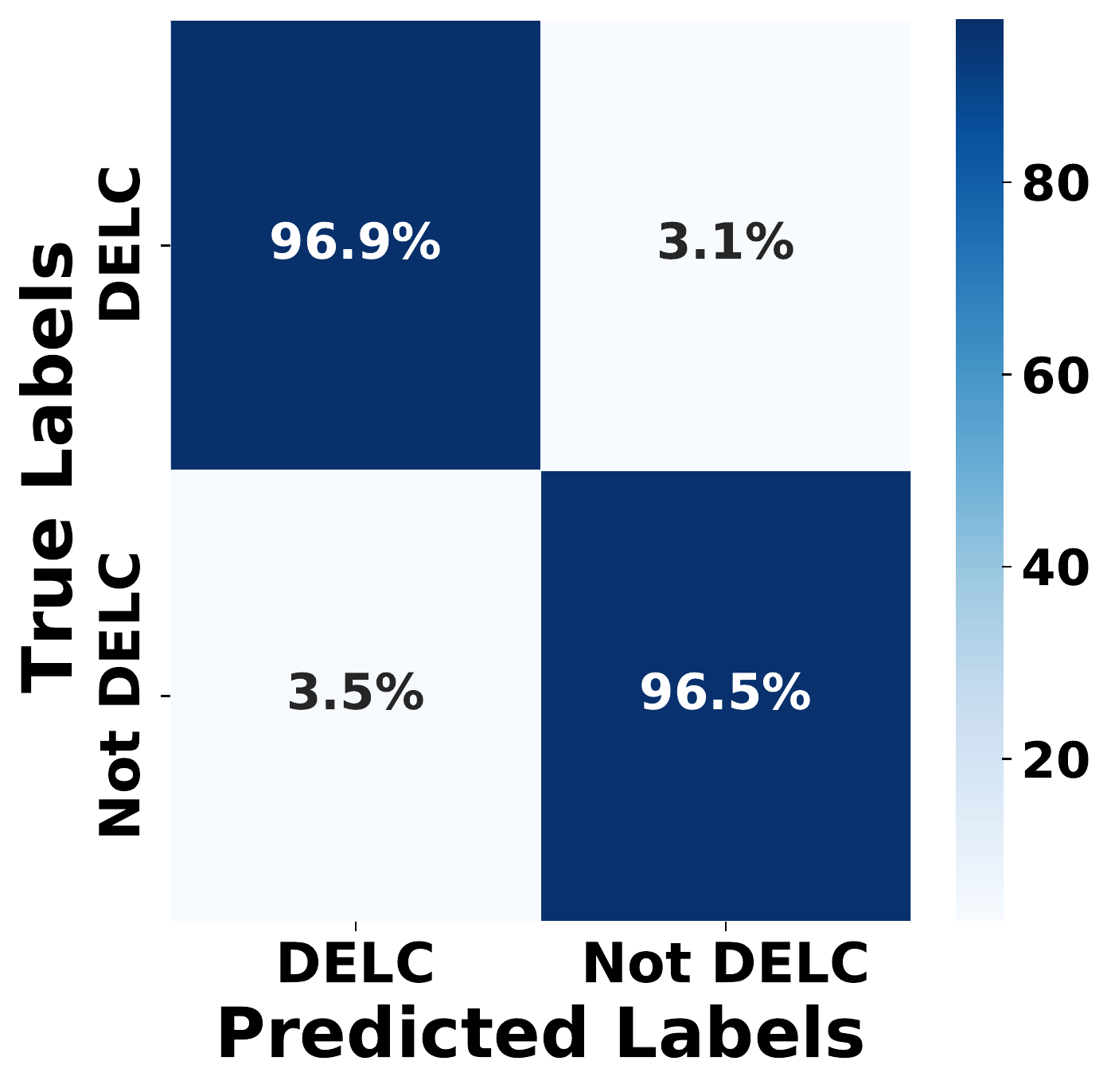}
     \caption{\textbf{MobileNet Confusion Matrix.}}\label{Fig:CF2}
   \end{minipage}
\end{figure*}

Table~\ref{tab:res} provides a summary of the results obtained of the considered backbones. Validation and test accuracy are provided to demonstrate the absence of overfitting. As expected, validation accuracy exceeds test accuracy by 1\% to 3\%. Adding additional layers or epochs did not improve these results. The table also displays the size of each backbone in millions of parameters. As can be seen, the backbone's size does not affect the model's robustness.

 Observing the table in detail, on the one side, it is evident that almost all pre-trained encoders perform remarkably detecting DELC. InceptionV3 achieves the best result ($97.7$\%), whereas DenseNet169 achieves the poorest ($88.1$\%).

Figure \ref{fig:relperf} better explores the relationship between backbones by displaying the relative performance and size. From the robustness standpoint, it is clear that InceptionV3 is more than 11\% superior to DenseNet169, the poorest backbone. VGG19 and VGG16 are the most prominent models in size, while the others fall between 20\% and 40\% below that size.

Figure \ref{fig:relperf} also highlights the relative significance of size when addressing the DELC issue. As can be seen, the smaller encoder within each family usually provides a superior performance. For instance, from the DenseNet models, DenseNet121 delivers a better outcome. The same is true for VGG models (VGG16$>$VGG19) and ResNet models (ResNet50$>$ResNet152$>$ResNet101).

The two top models support the most pertinent insight. An absolute difference of 1\%, but the MobileNet encoder is five and a half times smaller than the InceptionV3 encoder. In this regard, MobileNet performs three times faster than InceptionV3. However, a further error analysis of their performance is necessary to determine if their performance is robust in terms of balance predictions. In this regard, Figures \ref{Fig:CF1} and \ref{Fig:CF2} provide a closer examination of their confusion matrices. As can be seen, both models exhibit quite balanced performance, with the MobileNet encoder being the most promising due to its performance-size trade-off.

\section{\uppercase{Conclusions}}
\label{sec:con}

This paper presents a DELC classification analysis to determine whether or not an ear image contains Frank's Sign. Our research has shown a relationship between the size and performance of encoders. The reported experiments revealed that ear images could be used to sufficiently complete this task. Contrary to the literature, we have demonstrated that no further earlobe analysis is necessary. 

Unlike prior research, our study focuses on images in the wild. Our proposal includes the creation of a positive DELC subset and using an existing dataset (AWE dataset) to generate a negative DELC subset. Due to the shared acquisition method, both datasets can be considered valid for the task at hand: images from the Internet in unconstrained environments. In addition, data augmentation techniques are required during training due to subset-size limitations. 

We have effectively addressed this complex image scenario that requires interpretation based on encoder performance and size. The exploited encoder provides remarkable accuracy on the problem. However, we have shown that light-weight encoders usually perform better within the same backbones' family (ResNet, DenseNet, VGG). Moreover, due to their performance-size trade-off (-1\% performance, x3 times faster and x5.5 times lighter), we suggest the MobileNet as the most promising encoder.

This line of research presents numerous intriguing opportunities in healthcare scenarios. This proposal presents an opportunity to optimize pathways of diagnosis and prognosis and develop personalized treatment strategies by creating and utilizing larger datasets. Furthermore, analyzing pictures of earlobes for non-invasive DELC detection is among the most important applications. Besides, monitoring ears during aging is possible and may provide patient-specific insight into current health and alert medical staff to risk situations.

\section*{\uppercase{Acknowledgements}}

We want to acknowledge Dr. Cecilia Meiler-Rodríguez for her creative suggestions and inspiring ideas. This work is partially funded by the ULPGC under project ULPGC2018-08, the Spanish Ministry of Economy and Competitiveness (MINECO) under project RTI2018-093337-B-I00, the Spanish Ministry of Science and Innovation under projects PID2019- 107228RB-I00 and PID2021-122402OB-C22, and by the ACIISI-Gobierno de Canarias and European FEDER funds under projects ProID2020010024, ProID2021010012 and ULPGC Facilities Net and Grant  EIS 2021 04.


\bibliographystyle{apalike}
{\small

\end{document}